\documentclass[sigconf]{acmart} 
\renewcommand\footnotetextcopyrightpermission[1]{}
\AtBeginDocument{%
  }

\usepackage{afterpage}
\usepackage{lipsum}
\usepackage{placeins}
\usepackage{algorithm}
\usepackage{algpseudocode}

\acmDOI{}

\acmBooktitle{XAI-FIN-2025: International Joint Workshop on Explainable AI in Finance: Achieving Trustworthy Financial Decision-Making,
 November 15, 2025, Singapore}

\begin{document}

\title{Interpretable Model-Aware Counterfactual Explanations for Random Forest}

%
\author{Joshua S. Harvey}
\orcid{0000-0002-1342-9100}
\affiliation{%
  \institution{Prospect 33, LLC}
  \city{New York}
  \state{NY}
  \country{USA}
}
\email{joshua.harvey@prospect33.com}

\author{Guanchao Feng}
\affiliation{%
  \institution{BlackRock, Inc}
  \city{New York}
  \state{NY}
  \country{USA}
}
\email{guanchao.feng@blackrock.com}

\author{Sai Anusha Meesala}
\affiliation{%
 \institution{BlackRock, Inc}
  \city{Atlanta}
  \state{GA}
  \country{USA}
}
\email{saianusha.meesala@blackrock.com}

\author{Tina Zhao}
\affiliation{%
 \institution{BlackRock, Inc}
  \city{Atlanta}
  \state{GA}
  \country{USA}
}
\email{tina.zhao@blackrock.com}

\author{Dhagash Mehta}
\affiliation{%
 \institution{BlackRock, Inc}
  \city{New York}
  \state{NY}
  \country{USA}
}
\email{dhagash.mehta@blackrock.com}



\begin{abstract}
  Despite their enormous predictive power, machine learning models are often unsuitable for applications in regulated industries such as finance, due to their limited capacity to provide explanations. While model-agnostic frameworks such as Shapley values have proved to be convenient and popular, they rarely align with the kinds of causal explanations that are typically sought after. Counterfactual case-based explanations---where an individual is informed of which circumstances would need to be different to cause a change in outcome---may be more intuitive and actionable. However, finding appropriate counterfactual cases is an open challenge, as is interpreting which features are most critical for the change in outcome. Here, we pose the question of counterfactual search and interpretation in terms of similarity learning, exploiting the representation learned by the random forest predictive model itself. Once a counterfactual is found, the feature importance of the explanation is computed as a function of which random forest partitions are crossed in order to reach it from the original instance. We demonstrate this method on both the MNIST hand-drawn digit dataset and the German credit dataset, finding that it generates explanations that are sparser and more useful than Shapley values.
\end{abstract}




\maketitle
\pagestyle{plain}

\section{Introduction}

Machine learning (ML) models are increasingly being deployed in financial decision-making systems, ranging from credit scoring to asset management. However, the growing complexity of these models has led to concerns over a lack of transparency. In regulated environments such as finance, the ability to explain model decisions is not merely desirable—it is essential. Regulators, institutions, and end users alike require interpretability to ensure compliance, support trust, and provide accountability~\cite{arrieta2020explainable, guidotti2018survey}.

Explainability is particularly critical in credit scoring, where applicants are often entitled to know the reasons behind loan approvals or rejections~\cite{baesens2003benchmarking}. Similarly, in asset management, fund managers and compliance teams must be able to justify portfolio actions and assess risk exposure through interpretable reasoning~\cite{vellido2012making}. These demands highlight a key limitation of many contemporary ML approaches: while they may achieve high predictive accuracy, they often fail to provide intelligible insights into how or why decisions are made.

To address this, post hoc explanation tools have gained traction, with SHAP (Shapley Additive Explanations) emerging as one of the most widely adopted methods~\cite{lundberg2017unified}. SHAP attributes model predictions to input features by approximating Shapley values from cooperative game theory. Its formal guarantees and compatibility with a variety of model types, including tree ensembles, have made it a popular choice in financial applications~\cite{muller2020explaining}. Despite its success, SHAP has notable limitations. It typically assumes feature independence during explanation, which may lead to inaccurate attributions in domains with highly correlated variables—a common occurrence in financial data~\cite{kumari2023explainability}. Additionally, SHAP outputs are often abstract and numerical, lacking context in terms of real-world scenarios or comparable instances.

Counterfactual explanations have emerged as a promising alternative by answering the question: “How would input features need to change to produce a different prediction?” This approach aligns more directly with regulatory and human expectations of actionable insight \cite{barocas2020hidden}. Despite their appeal, generic counterfactual methods typically rely on surrogate models or optimization in artificial feature space—potentially producing unrealistic or uninterpretable instances that hurt fidelity \cite{laugel2019dangers}.

Tree-based models such as random forests (RFs) remain highly effective on tabular data, often outperforming deep learning methods in finance and other domains where structured data is predominant \cite{liu2015financial, biau2016random}. Their ensemble structure of decision trees naturally provides a notion of proximity between data points, which can be exploited to identify similar cases for explanation purposes. Unlike black-box neural networks, random forests offer intrinsic mechanisms such as out-of-bag proximity measures \cite{breiman2001random, breiman2017classification} that enable meaningful, model-aware similarity metrics without additional computational overhead. This unique feature makes random forests particularly well-suited for generating counterfactual explanations grounded in realistic, data-driven examples, enhancing interpretability in critical financial applications \cite{feng2024open, rossbach2018neural}.

In this work, we propose a complementary tree-based explainability approach grounded in case-based counterfactual reasoning. Our method is based on the framework we recently introduced~\cite{harvey2025explainable}, which is specifically designed for random forests. Instead of attributing predictions to marginal feature contributions, this approach identifies a counterfactual instance within the model feature space that yields a significantly different prediction. By constructing a trajectory from the original input to this counterfactual, the method traverses the decision boundaries of the forest in a series of interpretable steps. Each step reflects a minimal, model-aware change, and the cumulative feature transitions along this path form an intuitive and sparse explanation of the prediction.

This case-based approach has several advantages for financial modeling. First, it is inherently model-aware and leverages the discrete structure of tree ensembles to compute interpretable transitions without requiring surrogate models or approximation layers. Second, it produces explanations grounded in real or near-real examples, which can be more intuitive for stakeholders accustomed to precedent-based reasoning. For example, in credit risk modeling, rather than stating `Feature X contributed 40\% to the decision,' the explanation might read: 'This application differs from similar accepted applications in attributes A and B.' This frame of reference aligns better with human decision-making and regulatory expectations.

We present our method as a model-aware alternative to existing explainability frameworks, e.g., SHAP, particularly suited for settings where local fidelity and example-based reasoning are prioritized over additive feature attribution. To validate the approach, we first apply it to the Modified National Institute of Standards and Technology (MNIST) dataset \cite{lecun2010mnist}, a standard benchmark in explainability research, demonstrating the feasibility and faithfulness of counterfactual retrieval and trajectory construction. We then extend our analysis to structured financial data, demonstrating the methodology on the German Credit dataset \cite{statlog_(german_credit_data)_144} to evaluate its practical utility in credit scoring applications. By comparing the explanations generated by our case-based approach and SHAP, we assess their respective strengths in terms of interpretability, relevance for domain practitioners, and alignment with regulatory expectations. Our findings suggest that tree-based counterfactual explanations can offer more intuitive and actionable insights for decision-making in credit risk and asset management settings.

The remainder of this paper is organized as follows. In Section~\ref{sec:background}, we review related work on model explainability, with a particular focus on post hoc interpretation techniques and counterfactual reasoning in financial contexts. Section~\ref{sec:method} introduces our proposed method, detailing its algorithmic structure, counterfactual search strategy, and interpretability properties. Section~\ref{sec:experiments} presents empirical results on credit scoring and portfolio classification tasks, demonstrating the interpretive and practical value of our approach. Finally, Section~\ref{sec:conclusion} concludes with a discussion of implications, limitations, and directions for future research.

\section{Background}
\label{sec:background}
Post-hoc explainability methods can be broadly categorized into \emph{model-agnostic} and \emph{model-specific} approaches. Model-agnostic methods, such as LIME~\cite{ribeiro2016should} and SHAP~\cite{lundberg2017unified}, treat the model as a black box and approximate its local behavior through surrogate models or sampling. While these techniques are flexible and broadly applicable, their explanations may lack fidelity to the true decision process, especially in highly non-linear or structured models~\cite{slack2020fooling, aivodji2019fairwashing}.

In contrast, model-specific explainability methods are tailored to the internal structure of a given model class. For instance, decision trees and ensembles (e.g., random forests, gradient boosted trees) permit efficient introspection of splits, paths, and feature usage. This enables the construction of explanations that are both faithful and computationally tractable~\cite{lucic2022focus, tolomei2017interpretable}. Our proposed method follows this model-specific paradigm: it exploits the topology and proximity structure of tree-based models to generate case-based counterfactuals that are both plausible and aligned with the model’s learned decision boundaries.

Two complementary paradigms in post-hoc explainability are \textit{case-based} explanations and \textit{counterfactual} explanations. In case-based explainability, a model's prediction is interpreted by referencing similar instances (or \emph{cases}) from the training data. This approach reflects how human experts often justify decisions---by analogy to prior examples---and is common in domains such as medicine and law~\cite{kim2014bayesian, chen2019looks, keane2020good}. By identifying examples that share structural or semantic similarity with a given input, case-based methods can offer intuitive, example-driven justifications without requiring detailed knowledge of internal model parameters.

Counterfactual explanations, by contrast, aim to answer ``what-if'' questions: What is the smallest change to an input instance that would lead to a different prediction? This contrastive form of explanation aligns closely with how humans reason about causality and responsibility~\cite{byrne2019counterfactuals, miller2019explanation}, and has been widely adopted in legal and recourse-oriented contexts for its actionability and clarity~\cite{wachter2017counterfactual, fernandez2020explaining}. A key advantage of counterfactuals is that they provide users with specific, actionable changes they could make to obtain a desired outcome.

Recent research has explored how these two paradigms can be combined. \textit{Case-based counterfactual explanations} aim to generate contrastive explanations that are grounded in real or near-real instances---in contrast to optimization-based counterfactuals, which may yield implausible or out-of-distribution examples. Such hybrid approaches aim to preserve both interpretability and fidelity, and are particularly well-suited for model-specific structures, such as decision trees or ensembles, where notions of similarity and intervention can be precisely defined via model internals~\cite{keane2020good, lucic2022focus}.

Compared to model-agnostic methods such as SHAP~\cite{lundberg2017unified}, case-based counterfactual explanations are often simpler to communicate and better aligned with human intuitions of causality. While SHAP attributes importance scores to input features based on cooperative game theory, these scores can be difficult to interpret in isolation and may not correspond to actionable changes an end user can make. In contrast, case-based counterfactuals offer explanations in terms of \emph{concrete alternatives}---``what needs to change in this specific instance to alter the prediction''---which are more intuitive and user-centric. Furthermore, SHAP has been shown to \emph{struggle with reliability} in certain settings, at times performing no better than random baselines when evaluated on faithfulness or consistency~\cite{slack2020fooling}.

Our approach falls under the category of case-based counterfactual explainability, in which explanations are constructed by referencing real or closely related instances from the data. While counterfactual explanations generally aim to identify how inputs must change to yield a different model prediction, not all methods ensure that the resulting counterfactuals are plausible or data-supported. In contrast, case-based counterfactual methods explicitly ground the explanation in observed cases, offering intuitive and realistic reasoning paths. This contrasts with optimization-based approaches that may produce counterfactuals lying outside the data distribution. By combining case-based reasoning with counterfactual objectives, our method leverages both the interpretability of exemplar-based explanations and the actionable insights of counterfactual analysis—specifically tailored for tree-based models.

As pointed in \cite{dutta2022robust}, Tree-based ensembles pose additional challenges
in robust counterfactual generation, e.g., they have a non-smooth and non-differentiable objective function. While prior work such as \cite{dutta2022robust} focuses on generating robust counterfactuals for tree-based models—i.e.\ counterfactuals that remain valid under small model retraining or hyperparameter perturbations—our method takes a fundamentally different approach.
Instead of optimizing for robustness, we exploit the inherent random forest proximity measures (e.g., one version of it can be the frequency of co‑occurrence in leaf nodes) to identify similar training instances. We then compute the minimal number of decision path crossings required to move from the factual instance to a counterfactual instance. This yields case-based counterfactuals that are model-specific, faithful, and directly grounded in the machine learning model's internal geometry, rather than emphasizing cross-model consistency.

Particularly, random forests and other tree-based ensembles offer a unique advantage for case-based reasoning: they naturally induce a data-driven similarity metric based on how frequently two instances co-occur in the same leaf nodes across the ensemble. This notion, often referred to as \emph{tree proximity}~\cite{breiman2001random}, captures a form of contextual similarity that is model-aware, reflecting the decision boundaries learned by the ensemble rather than relying on arbitrary distance measures, which may be sensitive to feature scaling and normalization. In contrast to traditional instance-based approaches, which often use Euclidean or Mahalanobis distances in feature space, tree proximity is sensitive to feature interactions and nonlinearities implicit in the model. Leveraging this structure, our method generates counterfactual explanations that are not only actionable and sparse, but also anchored in regions of the input space deemed similar by the model itself. This contrasts with prior optimization-based approaches, such as Dutta et al.~\cite{dutta2022robust}, which emphasize robustness across retrained model instances but do not incorporate or exploit internal model geometry when defining similarity. By building on the random forest’s native proximity measure, our approach produces counterfactuals that are both interpretable and faithful to the underlying model, aligning with human intuitions of example-based reasoning.

RF-GAP (Random Forest Geometry- and Accuracy-Preserving)~\cite{rhodes2023geometry} proximity is a recently proposed framework that provides a refined notion of proximity for random forest models. Unlike classical tree-based proximity measures (e.g., the original proximity measure proposed in \cite{breiman2001random})—which count the frequency with which two instances fall into the same leaf node—RF-GAP leverages an explicit geometric interpretation of the ensemble’s decision space. It defines the proximity between instances based on their influence on the model's output, using kernel-based approximations to recover prediction behavior through a proximity-weighted sum (for regression) or majority vote (for classification). RF-GAP has been shown to closely match the out-of-bag (OOB) predictions of the original model, offering a more faithful and theoretically grounded characterization of instance similarity. This makes it particularly useful for tasks like counterfactual generation, instance-based explanations, and understanding local model behavior. In the literature, RF-GAP has been successfully adopted in various tasks, ranging from quantile regression \cite{li2024quantile}, enhancing local explainability \cite{rosaler2023towards} to quantifying outlier-ness \cite{desai2023quantifying} and open-set recognition \cite{feng2024open}. Therefore, in our work, we adopt RF-GAP for measuring similarity between data points. 






\section{Model Description}
\label{sec:method}
In this section, we describe our framework for generating model-specific counterfactual explanations using random forest models. Our approach is grounded in the observation that the structure of a trained random forest induces a natural proximity measure between data points, which we exploit to define plausible and model-faithful counterfactuals. We introduce the RF-GAP distance, a geodesic approximation of proximity based on tree traversal behavior, and demonstrate how it can be used to navigate from an input instance to a counterfactual example that alters the model's decision with minimal perturbation. Additionally, we describe how these counterfactual trajectories can be interpreted by tallying decision partition crossings, yielding sparse and transparent explanations that reflect the model’s internal reasoning. The subsections below detail the construction of RF-GAP distances, counterfactual path discovery, and the extraction of interpretable explanations from these paths.

\subsection{Counterfactual Explanations}

We begin with a general formulation of case-based counterfactual explanations in supervised learning. Let $f: \mathcal{X} \to \mathcal{Y}$ be a trained predictive model, where $x_i \in \mathcal{X}$ denotes an input and $y_i = f(x_i)$ its corresponding output. A counterfactual explanation for $x_i$ consists of an alternative input $x_c \in \mathcal{X}$ such that the model prediction at $x_c$ differs meaningfully from $y_i$:
\[
x_{c} \in \mathcal{X} \quad \text{such that} \quad y_{c} \neq y_i.
\]

To formalize “meaningfully different,” we introduce a utility function over outputs, $L_y: \mathcal{Y} \times \mathcal{Y} \to \mathbb{R}$, which quantifies the value or salience of a change in prediction:
\[
L_y(y_i, y_j) \;\longmapsto\; \text{utility gain from changing } y_i \text{ to } y_j.
\]
For regression or probabilistic classification settings, $L_y$ can often be instantiated as the difference $y_j - y_i$, although other domain-specific formulations are possible.

We then search for $x_{c}$, the closest instance in $\mathcal{X}$ that satisfies a required increase in utility, $\delta$. Defining a general distance measure over inputs, $d_x$:
\[
d_x : \mathcal{X} \times \mathcal{X} \;\longrightarrow\; [0,\infty),
\qquad
(x_i, x_j) \;\longmapsto\; d_x(x_i, x_j),
\]
the counterfactual explanation for instance $x_i$ is selected from a reference set $\{x_j\}_{j=1}^n$ by solving
\begin{equation}\label{eq:implicit}
c = \mathop{\arg\min}_{j\in\{1,\dots,n\}} \; d_x(x_i,x_j)
\quad\text{subject to}\quad
L_y(y_i,y_j) > \delta,
\end{equation}
where $\delta$ is the minimal surpassed increase in utility. This formulation casts counterfactual search as a constrained nearest-neighbor retrieval problem.


\subsection{Random Forest–Based Counterfactuals}

Once the counterfactual case, $x_c$, is identified, it can simply be reported to a user in terms of its feature values, and how they differ with respect to $x_i$. Unless many feature values are exactly equal (giving a sparse explanation), this does not explain which features are most important in giving rise to the change in the utility function. To address this, we can relate feature value changes with the binary decisions (partitions) of the random forest model. Drawing a line segment between $x_i$ to $x_c$ in input space, we tally the random forest partition intersections, as described in \cite{harvey2025explainable}. This renders a counterfactual in terms of a signed partition tally, allowing features to be expressed in terms of their salience to the model, and whether the instance's values should be increased or decreased to alter the prediction.

Similarly, we can re-frame the search for the counterfactual in terms of similarity from the `perspective' of the random forest model, by using the geometry- and accuracy-preserving random forest proximity (RF-GAP, \cite{rhodes}) to locate nearby instances:

\begin{equation} \label{GAPprox}
p^{GAP}_{i,j} = \frac{1}{|S_{i}|} \sum_{t \in S_{i}} \frac{c_j(t)\,\mathbb{I}[j \in J_{i}(t)]}{|M_{i}(t)|},
\end{equation}
where:
\begin{itemize}
    \item $S_i$ is the set of trees for which $x_i$ is out-of-bag;
    \item $M_i(t)$ is the multiset of training instances in the same leaf as $x_i$ in tree $t$;
    \item $J_i(t)$ is the deduplicated set of those instances (i.e., the support of $M_i(t)$);
    \item $c_j(t)$ is the number of times instance $j$ appears in the bootstrap sample of tree $t$.
\end{itemize}

This leads to the following symmetric RF-GAP distance:
\begin{equation} \label{GAPdist}
d_x := d^{GAP}_{i,j} =
\begin{cases}
0, & \text{if } i = j \\
(0.5 \cdot (p^{GAP}_{i,j} + p^{GAP}_{j,i}))^{-1}, & \text{otherwise}.
\end{cases}
\end{equation}

Using random forest distances, $d_x^{GAP}$, ensures that counterfactual cases are found that minimally differ from the perspective of the model, but result in maximal changes to the utility function. When combined with the rendering of counterfactual cases as partition tallies, this should result in sparser explanations.


\subsection{Counterfactual Trajectories}

In the limiting case where $\delta=0$, this approach is likely to find a very close counterfactual case, but whose corresponding output only weakly differs from the original instance. Such counterfactual cases may therefore have weak explanatory power, and be highly susceptible to noise. Increasing $\delta$, such as by requiring a `flip' in the predicted class output, will provide stronger counterfactual cases. However, this approach may fail to find sufficiently similar cases to the original instance---the counterfactual may be starkly different from the instance along multiple features and, therefore, be of little use. This is especially the case in high-dimensional data that exhibits well-defined manifold (correlation) structure, for which larger values of $d_x$ become uninformative. When data is tightly constrained to a manifold, random forest partitions may be located only where the measurement space is densely populated with data, particularly for models of greater depth. Relating large jumps to random forest partitions may not yield informative explanations in this case, if such jumps span regions of measurement space that are mostly empty.

To solve this issue, a counterfactual trajectory may be identified that charts a series of local steps through the data, optimizing the utility function as it evolves (`hill climbing'). By exploiting the local structure of the data through a series of geodesics, this approach ensures that counterfactual explanations are framed according to realizable and observed changes in input features, thereby preventing discontinuities that would require changing a large number of features. A trajectory constructed in this manner allows us to optimize the trade-off between the interpretability (sparseness) of counterfactual explanations on the one hand, and their explanatory power on the other. Trajectory steps may be computed until convergence (i.e., the utility function reaches its global maximum), as in Algorithm 1, or until a criterion is reached (such as a class flip occurring). We should note this is a different usage of the term from prior literature, where it denotes counterfactual explanations comprising a series of sequential decisions or actions \cite{clark2024trace}.

\begin{algorithm}[H]
\label{alg:trajectory}
\caption{Counterfactual Trajectory}
\begin{algorithmic}[1]

\Require
  $\mathbf{X} = \{\mathbf{x}_1,\dots,\mathbf{x}_n\}$ \Comment{Data points in $\mathbb{R}^d$} \\
  $D: \{\mathbf{x}_i,\mathbf{x}_j\} \mapsto D_{ij}^{GAP}$ \Comment{Random forest distances} \\
  $U: \{\mathbf{x}_i\} \mapsto \mathbb{R}$ \Comment{Utility function} \\
  $t_0$ \Comment{Index of starting point}
\Ensure
  Trajectory indices $(t_0, t_1, \dots, t_T)$ until no further increase

\State $\text{traj} \gets [t_0]$
\State $i \gets t_0$
\Repeat
  \State $\mathcal{C} \gets \{\,j \mid U(\mathbf{x}_j) > U(\mathbf{x}_i)\}$
  \If{$\mathcal{C} = \emptyset$}
    \State \Return traj
  \EndIf
  \State $i \gets \arg\min_{j \in \mathcal{C}} D_{ij}$
  \State Append $i$ to traj
\Until{$\mathcal{C} = \emptyset$}
\end{algorithmic}
\end{algorithm}

\section{Experiments and Results}
\label{sec:experiments}

To evaluate the practical utility and interpretability of our counterfactual trajectory framework, we conduct experiments on two widely used benchmark datasets: the MNIST image dataset and the German Credit dataset. These datasets represent two distinct domains—vision and structured tabular data—and enable us to assess the method's generalizability and fidelity across modalities. We focus our evaluation on the interpretability, sparsity, and local fidelity of the generated explanations. The following subsections present the results for each dataset in turn.



\subsection{MNIST dataset}
We begin by validating our method on the MNIST dataset \cite{lecun2010mnist}, a widely used benchmark in the machine learning community for evaluating classification models and explainability techniques. MNIST consists of 70{,}000 grayscale images of handwritten digits (0–9), each of size $28 \times 28$ pixels, split into 60{,}000 training and 10{,}000 test examples. Its simplicity, balanced class distribution, and intuitive visual structure make it an ideal setting for assessing whether model explanations align with human perception. In particular, MNIST allows for visual validation of counterfactual directions and attributions—critical for evaluating the coherence and sparsity of example-based explanations.

This controlled experimental setting also enables transparent comparisons with established explainability frameworks such as SHAP \cite{lundberg2017unified}, LIME \cite{ribeiro2016should}, TreeInterpreter \cite{sharma2020evaluating}, and Accumulated Local Effects \cite{apley2020visualizing}. By grounding our analysis in this benchmark, we aim to demonstrate the plausibility and effectiveness of our method in a well-understood, high-signal regime before turning to real-world financial datasets.

\begin{figure*}[h]
  \centering
  \includegraphics[width=\linewidth]{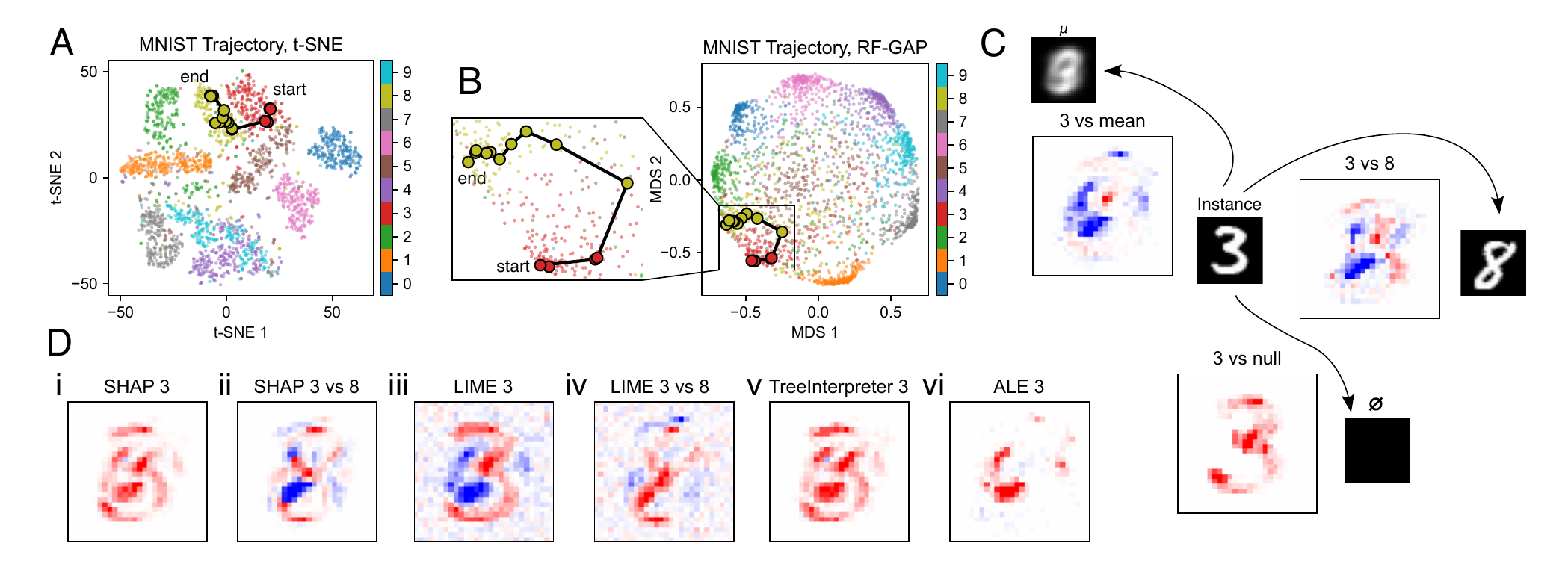}
  \caption{Counterfactual trajectory explanations on the MNIST dataset. a) A counterfactual trajectory shown for a digit `3' vs the label `8', plotted in a \textit{t}-SNE embedding of the dataset. b) The same trajectory plotted in an MDS embedding of RF-GAP distances. c) Visualization of different counterfactual explanations that may be obtained by tallying random forest partition crossings. d) Comparison with other common ML explanation frameworks.}
  \label{fig:fig1}
\end{figure*}

To this end, we trained a multiclass random forest classifier on the MNIST dataset (containing 1,000 trees, with a maximum depth of 5 splits, test-set accuracy = $86\%$) and applied our explainability framework to generate counterfactual trajectories. For example, to explain why a digit labeled `3' was not predicted as an `8', we identified a path through input space that maximizes the predicted probability of the `8' class while remaining locally faithful to the model's internal structure. Figure~\ref{fig:fig1}.a displays this counterfactual trajectory in a \textit{t}-SNE embedding of the dataset, where points are colored by their true class. Figure~\ref{fig:fig1}.b shows the same trajectory embedded via multidimensional scaling (MDS) \cite{hout2013multidimensional, carroll1998multidimensional} applied to RF-GAP distances—distances computed from the random forest’s internal structure—highlighting the local proximity relationships encoded by the model.

As illustrated in the inset of Figure~\ref{fig:fig1}.b, the trajectory captures localized case-based counterfactual shifts that are minimally disruptive in the model’s representation space. These paths are then used to generate interpretable explanations. Figure~\ref{fig:fig1}.c visualizes these explanations by tallying the random forest partition crossings encountered along the trajectory. The plot labeled `3 vs null' shows the partition tallies for the `3' instance vs a null input of zero values. Red pixels indicate where random forest partitions have been crossed for a given feature in the positive direction, to reach the instance from the counterfactual. As such, they explain which regions of the image the random forest model is sensitive to, vs an empty set. The plot labeled ‘3 vs 8’ corresponds to the counterfactual trajectory in Figure~\ref{fig:fig1}.b; red pixels indicate input features that would drive predictions toward `3’ (positive-signed random forest partition tallies integrated over the trajectory from counterfactual to instance), and blue pixels indicate features that would promote an `8’ prediction. We also show the random forest partition tally explanation for the `3 vs mean’ counterfactual.

Figure~\ref{fig:fig1}.d compares the outputs of our method with those produced by SHAP, LIME, TreeInterpreter, and ALE. Note that the SHAP explanation for the instance's `3' class prediction value is not easy to interpret (fig.~\ref{fig:fig1}.d.i), as it conflates regions of the image where high input values drive the `3' class prediction, and where low input values drive the `3' class prediction. In that sense, it is comparable to an unsigned combination of the `3 vs null' and `3 vs mean' partition tallies shown in Figure~\ref{fig:fig1}.c. The LIME explanation (fig.~\ref{fig:fig1}.d.iii) retains the polarity of whether pixels drive a `3' class prediction to their values being high (red) or low (blue), but exhibits remarkably low sparsity, even in empty regions of the input. While SHAP produces sparser results than LIME, our method achieves even greater sparsity by aggregating minimal, model-aligned changes across trajectory steps (RF-partition sparsity $=0.6253 \pm 0.015$, SHAP sparsity $=0.3839 \pm 0.0114$, $p<0.001$). These findings underscore the effectiveness of our case-based counterfactual approach in yielding concise, actionable, and model-faithful explanations.

\subsection{German Credit dataset}

\begin{figure*}[h]
  \centering
  \includegraphics[width=\linewidth]{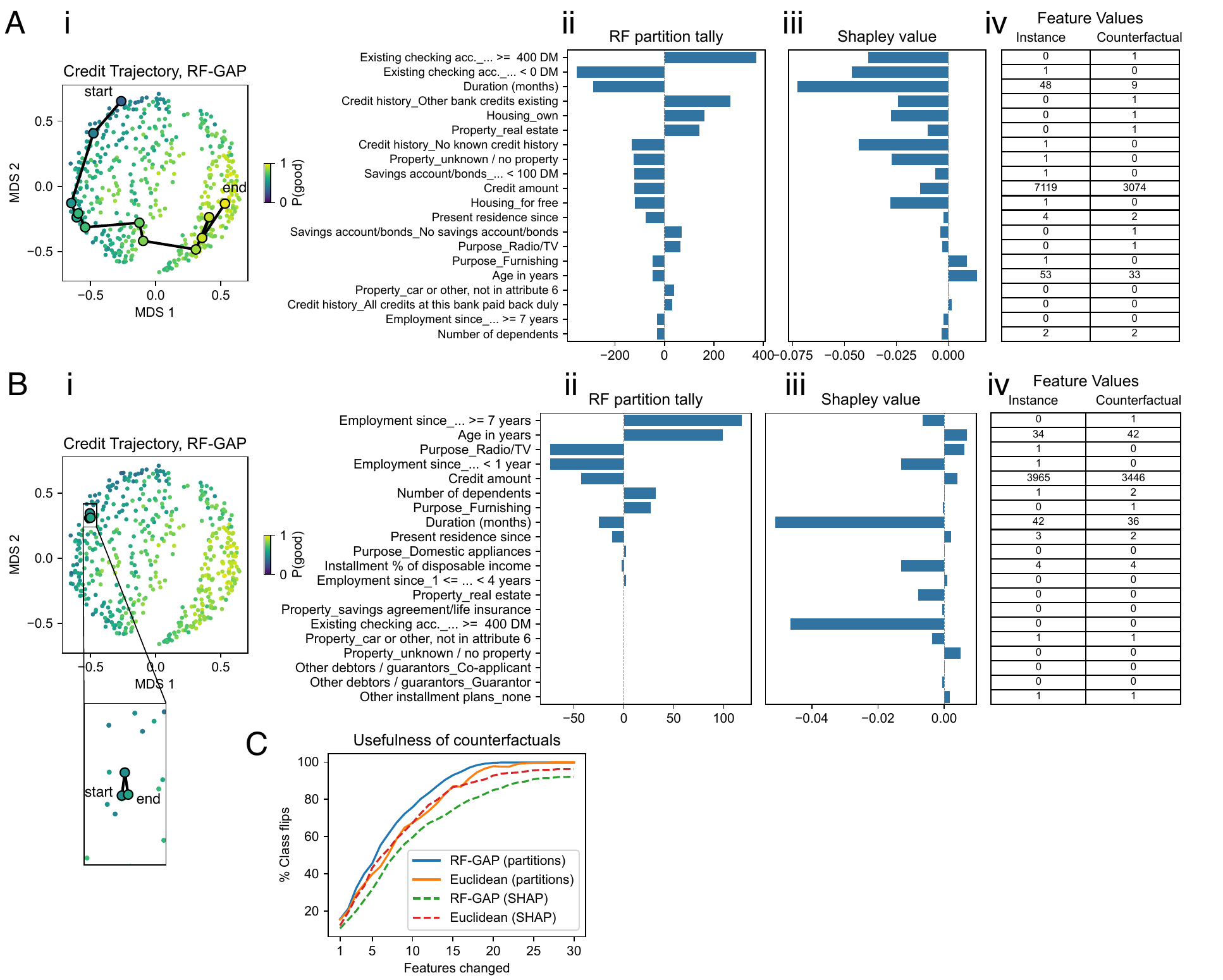}
  \caption{Explanations for the German credit dataset. a) The counterfactual explanation for the instance in the test set with the lowest credit worthiness vs the instance with the highest credit worthiness. i) The counterfactual trajectory plotted in an embedding of the dataset using the RF-GAP distances from the predictive model. ii) The random forest partition tally integrated along the counterfactual trajectory, sorted by absolute value for the top 20 features. iii) The Shapley values for the same features in (ii). The values of the instance and its counterfactual for those features. b) A counterfactual relationship for an instance just shy of `good' credit worthiness ($P(good)=0.49$) vs its closest `class flip' ($P(good)=0.58$). c) Usefulness of counterfactuals, evaluated as frequency of class flips against number of features changed to that of an instance's counterfactual, for every instance in the test set ($n=500$).}
  \label{fig:fig2}
\end{figure*}

A frequent use-case for counterfactual explanations, or \textit{algorithmic recourse}, is for models predicting credit worthiness. Here, we demonstrate our method for random forest explainability on the German credit dataset \cite{statlog_(german_credit_data)_144}, which classifies people described by a set of attributes as good or bad credit risks. The dataset contains 1,000 entries measured across 21 features, which we expand to 61 after one-hot-encoding of categorical variables. As before, we train a random forest classifier with 1,000 trees and a maximum depth of 5, on a training set containing 50\% of the data (test-set accuracy $=74\%$). Figure \ref{fig:fig2}.a shows a counterfactual explanation for the instance in the test set with the worst credit worthiness, vs the instance with the best credit worthiness, with Figure \ref{fig:fig2}.a.i showing its counterfactual trajectory in an RF-GAP embedding of the dataset.

The tallies of random forest partition crossings along the trajectory are shown in Figure \ref{fig:fig2}.a.ii, with features sorted according to the top 20 absolute tally values. Shapley values are also shown for each feature (fig. \ref{fig:fig2}.a.iii), as well as the feature values for both the instance and the final point in the counterfactual trajectory (fig. \ref{fig:fig2}.a.iv). While both RF partition crossing and SHAP explanations can take positive or negative values for features, the interpretation of the sign of explanation differs. For RF partition crossings, a positive tally indicates that the route from instance to counterfactual entails increasing a feature value, such as from a low numerical value to a high value, or from 0 to 1 in the case of one-hot-encoded features. For the example in Figure \ref{fig:fig2}.a, the feature with the most tallies is `Existing checking account >= 400 DM'---whether the customer has an existing checking account with more than 400 Deutschemark. As may be expected, the instance with poor credit worthiness does not meet this criterion (0), the counterfactual does (1), and along the counterfactual trajectory a large number of RF partitions on this feature are crossed in the positive direction. The SHAP value for this feature, however, is negative, because the inclusion of this feature (over all potential combinations of other features) causes an overall reduction in the predicted outcome---the predicted probability of good credit worthiness. The fact that the features with the greatest (absolute) RF partition tallies also have strongly negative Shapley values indicates good agreement between the two explanatory methods, which is not surprising for the particular counterfactual relationship (the worst vs best credit worthiness instances in the test set).

Another counterfactual explanation is shown in Figure \ref{fig:fig2}.b, for an instance just shy of having `good' credit worthiness ($P(good)=0.49$), vs its nearest `class flip', which is classified as having `good' credit worthiness ($P(good)=0.58$). Of note is the sparsity of explanation, with only 12 of the 61 features taking non-zero values for RF partition tallies (fig. \ref{fig:fig2}.b.ii). Also of interest is that the feature with the second highest tally, `Age in years', actually has a positive Shapley value, indicating that this feature has a positive influence on the credit worthiness of the instance. However, changing only this feature to take the value of its counterfactual (from 34 to 42, as shown in fig. \ref{fig:fig2}.b.iii), is sufficient to flip its class from bad credit worthiness to good credit worthiness ($P(good)=0.52$). Although it may be common practice to interpret counterfactual cases through the use of Shapley values, so as to infer feature importance towards actionable recommendations, this is a clear example of where doing so may be counterproductive.


To test the usefulness of our method vs Shapley values for interpreting the feature importance of counterfactual relationships, we performed an experiment evaluating which method is more successful at identifying the features of an instance that, when changed to take the value of its counterfactual, results in a class flip. For every instance in the test set ($n=500$), we first identified its counterfactual case, locating the closest point with a flipped class as measured according to either Euclidean or RF-GAP distances. The feature importance of the counterfactual relationship was then rendered according to either the tally of random forest partitions crossed to reach it from the original instance (`(partitions)'), or the Shapley values of the original instance (`(SHAP)'). Note that in the case of the former, feature importance depends on both the original and counterfactual points, while the SHAP framework feature importance depends only on the original. The original point was then perturbed, taking on the top-\textit{k} most important features of the counterfactual according to each method. Across the whole test set, we then evaluated if such perturbations resulted in a class flip for each value of \textit{k}. As shown in Figure \ref{fig:fig2}.c, identifying counterfactual cases by random forest distances, and interpreting their feature importance according to partition crossings (`RF-GAP (partitions)') is the dominant strategy, as perturbations according to this approach are the most likely to result in class flips for any number of features perturbed.

\section{Conclusion}
\label{sec:conclusion}

In this work, we introduced a model-specific framework for generating interpretable counterfactual explanations tailored to random forest models. By leveraging the internal representation of the data learned by the model---operationalized through RF-GAP distances---we identified counterfactual instances that are both plausible and minimally divergent from the original input. These counterfactuals were further interpreted through a tallying mechanism over decision tree partition crossings, providing sparse, feature-level explanations grounded in the model's own structure and logic.

Our empirical evaluation on both visual (MNIST) and structured tabular (German Credit) datasets highlights several strengths of the approach. First, the use of RF-derived distances ensures that counterfactuals remain locally faithful to the model’s behavior, maintaining semantic coherence while modifying predictions. Second, by tracing the specific decision boundaries crossed along a counterfactual trajectory, we obtain sparse and actionable feature attributions that are often more interpretable than traditional additive methods like SHAP or LIME. In particular, our method yields counterfactuals that offer higher-quality recourse---interventions that not only change the model’s decision but also align with intuitive, human-understandable changes.

These findings demonstrate the potential of tree-based, case-driven explainability as a viable and interpretable alternative to more abstract attribution methods, particularly in high-stakes domains such as credit scoring and asset management. Unlike purely statistical decompositions, our method emphasizes example-based reasoning, which is often more aligned with how domain experts and regulators evaluate decision-making.

Future work will explore extending this framework to other ensemble models and improving computational efficiency for large-scale applications. We also intend to investigate hybrid approaches that integrate our case-based logic with feature attribution techniques to provide layered explanations. Finally, we see significant potential in applying this method to other regulated domains such as healthcare, where fidelity, transparency, and actionable insight are paramount.

\begin{acks}
The views expressed here are those of the authors alone and not of BlackRock, Inc.
\end{acks}




\bibliographystyle{ACM-Reference-Format}
\bibliography{references.bib}

\end{document}